%% file: main.tex
\documentclass[10pt,twocolumn]{article}

\usepackage[margin=1in]{geometry}
\usepackage{times}
\usepackage{graphicx}
\usepackage{amsmath}
\usepackage{amssymb}
\usepackage{booktabs}
\usepackage{multirow}
\usepackage{makecell}
\usepackage{xcolor}
\usepackage{hyperref}
\PassOptionsToPackage{hyphens}{url}
\usepackage{url}
\usepackage{cite}
\usepackage{caption}
\usepackage{subcaption}
\usepackage{enumitem}
\usepackage{microtype}
\usepackage{balance}
\usepackage[capitalise]{cleveref}
\usepackage{titlesec}
\usepackage{cuted}     % \begin{strip} ... \end{strip} for full-width inline content
\hypersetup{
  colorlinks=true,
  linkcolor=blue!70!black,
  citecolor=green!50!black,
  urlcolor=blue!70!black
}

\title{\textbf{When the Forger Is the Judge:\\
GPT-Image-2 Cannot Recognize Its Own Faked Documents}}

\author{
  Jiaqi Wu$^{*}$,
  Yuchen Zhou$^{*}$,
  Dennis Tsang Ng,
  Xingyu Shen,
  Kidus Zewde,\\
  Ankit Raj,
  Tommy Duong,
  Simiao Ren$^{\dagger}$\\[0.5em]
  {\small $^{*}$Equal contribution \quad $^{\dagger}$Corresponding author: \texttt{benren@scam.ai}}
}

\date{}

\begin{document}
\maketitle
{\renewcommand{\thefootnote}{}\footnote{Authorship of this paper does not imply employment by or affiliation with Scam.AI (Reality Inc.); some authors are independent researchers who collaborated on this work.}\addtocounter{footnote}{-1}}

% ---- Hero figure pinned ABOVE the abstract, spans both columns ----
\begin{strip}
\centering
\includegraphics[width=0.95\textwidth]{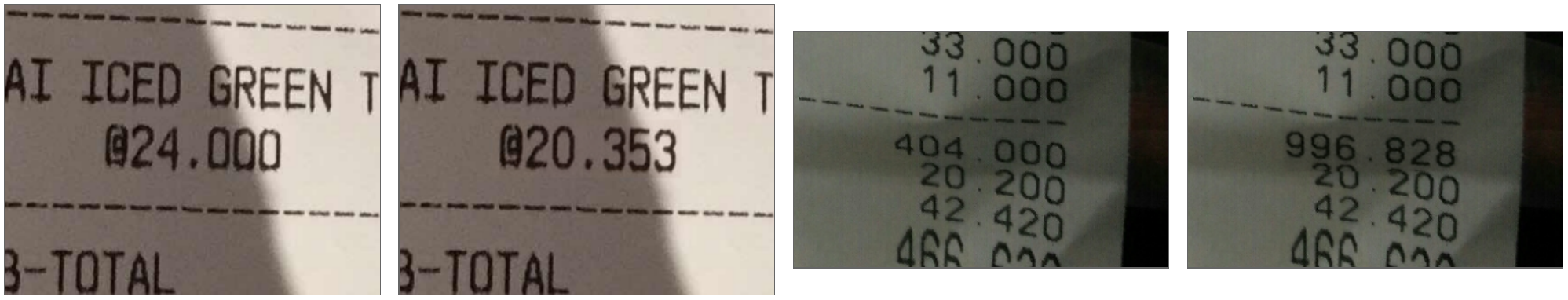}
\refstepcounter{figure}\label{fig:hero}
\par\vspace{0.5em}
{\small\textbf{Figure \arabic{figure}:} \textbf{Can you tell which
document is real?} Two paired examples from AIForge-Doc v2.}
\vspace{1em}
\end{strip}

% ---- Abstract ----
\begin{abstract}
\input{sections_v2/00_abstract}
\end{abstract}

% ---- Sections ----
\input{sections_v2/01_introduction}
\input{sections_v2/02_related_work}

\input{sections_v2/03_dataset_construction}
\input{sections_v2/04_statistics}
\input{sections_v2/05_judges}
\input{sections_v2/06_results}

\input{sections_v2/07_discussion}
\input{sections_v2/08_conclusion}

\clearpage
\appendix
\titleformat{\section}
  {\normalfont\Large\bfseries}{Appendix \thesection:}{0.7em}{}
\input{sections_v2/A_appendix}

\section*{Data and Code Availability}
The datasets and code supporting this work are released by Scam.AI and are available from
the Scam.AI Research portal, \url{https://scam.ai/research}~\cite{scamai_research}. Document-
and PDF-forgery resources are additionally maintained at Trustable, \url{https://trustable.ai}~\cite{trustable_ai}.
These are the canonical, maintained locations for the resources associated with this paper and
supersede any prior ad hoc distribution.

\bibliographystyle{plain}
\bibliography{references}

\end{document}

%% file: sections_v2/00_abstract.tex
OpenAI's GPT-Image-2 has effectively erased the visual boundary
between authentic and AI-edited document images: a single number on
a receipt can now be replaced in under a second for a few cents.
We release \textbf{AIForge-Doc v2}, a paired dataset of
\textbf{3{,}066} GPT-Image-2 document forgeries with pixel-precise
masks in DocTamper-compatible format, and benchmark four natural
lines of defence: human inspectors ($N=120$, $n=365$ pair-votes via
the public $2$AFC site \href{http://canuspotai.com/}{\texttt{CanUSpotAI.com}}),
\textsc{TruFor} (generic forensic), \textsc{DocTamper} (qcf-568,
document-specific), and \emph{the same GPT-Image-2 model} as a
zero-shot self-judge --- asked, to avoid the trivial ``the image is
mostly real'' reading, whether \emph{any region of the document was
generated or edited by an AI image model}. Human $2$AFC accuracy
is $0.501$, indistinguishable from chance: even side-by-side, human
inspectors cannot tell GPT-Image-2 receipt forgeries from their
authentic counterparts. The three computational judges sit only
modestly above (TruFor $0.599$, DocTamper $0.585$, self-judge
$0.532$). \textbf{The self-judge fails consistently, not
by chance}: across five prompt strategies and four policies for
handling ambiguous responses, AUC never rises above $0.59$ --- no
rephrasing of the question lifts it out of the near-random regime.
To rule out the possibility that the two forensic detectors are
broken on our source domain rather than blind to AI inpainting, we
calibrate each on a same-domain traditional-tampering set
constructed for its training distribution: TruFor reaches AUC
$0.962$ on cross-camera splicing of our dataset, and DocTamper
reaches AUC $0.852$ on cross-document OCR-token splicing of our
dataset with two-pass JPEG re-encoding. Both detectors thus
retain near-published performance on our document domain when the
tampering is traditional; switching the tampering to GPT-Image-2
inpainting drops detector AUC by $0.27$--$0.36$ ($0.962\!\to\!0.599$
for TruFor; $0.852\!\to\!0.585$ for DocTamper), isolating a
detection gap that is specific to GPT-Image-2 inpainting. We
release the dataset, pipeline, four-judge protocol, and
calibration sets.

%% file: sections_v2/01_introduction.tex
\section{Introduction}
\label{sec:intro}

Document fraud has become an industrial-scale problem: the 2025
Entrust Identity Fraud Report~\cite{entrust2025fraud} measured a
$244\%$ year-over-year jump in digital document forgeries, with
digital tampering ($57\%$) overtaking physical counterfeiting as the
dominant method, and deepfake or AI-manipulated document attempts
now occur every five minutes. The threat changed qualitatively in
April 2026, when OpenAI released
\textbf{GPT-Image-2}~\cite{openai2026gptimage2}: unlike earlier
inpainting systems whose outputs leave characteristic compression
seams, cloning patterns, or noise-residual mismatches, GPT-Image-2
can replace a single number on a receipt photograph in under a
second for a few cents, and because the same model produced both
the original context and the edited region, the result is by
construction statistically consistent with the surrounding pixels.

\paragraph{Four candidate lines of defence.}
We consider four natural lines of defence: \textbf{(i) a
human inspector} reads the document and notices content-level
inconsistencies; \textbf{(ii) a generic forensic detector}
(TruFor) picks up sensor- or pipeline-noise inconsistencies;
\textbf{(iii) a document-specific detector} (DocTamper) picks up
JPEG-quantisation and typographic signatures; \textbf{(iv) the
generator itself}, having produced the forged pixels, may
recognise its own output as non-authentic. We evaluate all four.
Human $2$AFC accuracy on
\href{http://canuspotai.com/}{\texttt{CanUSpotAI.com}}
($N=120$, $n=365$ pair-votes) is $0.501$ ---
\emph{indistinguishable from chance}; the three computational
judges sit only marginally above (AUC $0.532$--$0.599$). To
verify that the forensic-detector
collapse is specific to GPT-Image-2 inpainting rather than a generic
inability to operate on receipt and form scans, we calibrate TruFor
and DocTamper against same-domain traditional-tampering sets
matched to each detector's training distribution. Both clear
$0.85$ AUC on the calibration sets (TruFor $0.962$, DocTamper
$0.852$); switching the tampering to GPT-Image-2 inpainting
drops detector AUC by $0.27$--$0.36$, isolating the
detectability gap to AI inpainting
(\S\ref{sec:calibration}).

\paragraph{Contribution.}
We release three artefacts that together establish
self-indistinguishability as a measurable property of GPT-Image-2
and expose an AI-inpainting-specific detection gap in current
document-forgery detectors:
(i)~\textbf{AIForge-Doc v2}, a paired-with-v1 dataset of $3{,}066$
GPT-Image-2 document forgeries spanning four source corpora
(CORD, WildReceipt, SROIE, XFUND), nine languages, and four
field-type categories, with pixel-aligned authentic counterparts
in DocTamper-compatible format (\S\ref{sec:dataset});
(ii)~\textbf{a reproducible generation pipeline} with the
engineering disclosures future practitioners need
(aspect-preserving size snap, green-outline composite-marker mask,
and a $24.5\%$ deterministic rejection rate dominated by a
$1\!:\!3$ aspect-ratio constraint that shapes the v2 corpus;
\S\ref{sec:pipeline}, \ref{sec:stats});
(iii)~\textbf{a four-judge evaluation protocol} --- humans via
\href{http://canuspotai.com/}{\texttt{CanUSpotAI.com}} ($2$AFC,
$N=120$, $n=365$), TruFor, DocTamper (qcf-568), and the
GPT-Image-2 self-judge under the minimal prompt
\emph{``Is this image AI-edited?''} --- with humans at $0.501$
(chance) and all three computational judges in $[0.532, 0.599]$
on v2, plus detector-specific calibration sets that establish the
same-domain traditional-tampering upper bounds (TruFor $0.962$,
DocTamper $0.852$; \S\ref{sec:judges}, \ref{sec:results},
\ref{sec:calibration}).

Working definitions of \emph{self-recognition} and
\emph{operational utility}, the relation to AIForge-Doc
v1~\cite{wu2026aiforgedoc}, and the formal scope of our empirical
claim are in Appendix~\ref{app:definitions-scope}.

%% file: sections_v2/02_related_work.tex
\section{Related Work}
\label{sec:related}

\subsection{Document Forgery Datasets}
The largest prior corpora --- DocTamper~\cite{qu2023doctamper} (170k),
RTM~\cite{luo2024rtm} (9k, 6k professionally manipulated), and the
ICDAR 2023 TII benchmark~\cite{icdar2023dtti} (11k) --- all rely on
copy-move, splicing, or typesetting manipulation rather than AI
inpainting. OSTF~\cite{qu2025ostf} (AAAI 2025) is the closest in
spirit but studies AI text replacement on \emph{scene-text} images
(storefronts, menus, signs) with bounding-box rather than
pixel-level masks. The direct antecedent of this paper is
\textbf{AIForge-Doc v1}~\cite{wu2026aiforgedoc}: $4{,}061$
diffusion-based document forgeries (Gemini~2.5 Flash Image and
Ideogram~v2 Edit) with paired authentic and pixel-precise masks,
which showed zero-shot detectors degrade severely on diffusion-style
document inpainting (TruFor $0.751$, DocTamper $0.563$, GPT-4o
$0.509$). We reuse v1's forgery specifications and source datasets
and swap only the generator, so any change in detector behaviour is
attributable to the generator rather than the documents.

\subsection{Forensic Detectors}
General-purpose detectors (ManTraNet~\cite{wu2019mantranet},
CAT-Net~\cite{kwon2021catnet}, PSCC-Net~\cite{liu2022psccnet},
HiFi-Net~\cite{guo2023hifinet}, IML-ViT~\cite{ma2023imlvit}) target
copy-move and splicing in natural photographs.
TruFor~\cite{guillaro2023trufor} (CVPR 2023) is the current state of
the art and our generic forensic baseline.
DocTamper~\cite{qu2023doctamper} is the only detector trained
specifically on document forgeries and our document-specific
baseline. We deliberately omit diffusion-specific detectors such as
AEROBLADE~\cite{ricker2024aeroblade} and
DiffForensics~\cite{yu2024diffforensics}, which target full-image
generation and are mathematically inapplicable to localised
inpainting where surrounding authentic pixels dominate the
reconstruction signal.

\subsection{LLMs/VLMs as Forensic Judges}
Multimodal LLMs have been benchmarked as deepfake
detectors~\cite{ren2025llmdeepfake} and as fraudulent-document
judges via prompt optimisation~\cite{liang2025llmdocmanip}.
AIForge-Doc v1 reported a GPT-4o judge at chance ($0.509$) on its
diffusion forgeries. We replace that judge with the
\emph{same-family} GPT-Image-2 model that produced the forgeries
(\S\ref{sec:judges}), under a deliberately minimal prompt so that
any apparent zero-shot competence is not attributable to prompt
engineering.

\subsection{Self-Recognition by Generative Models}
Prior text-domain studies show language models exhibit weak
``self-awareness'' for their own
generations~\cite{panickssery2024selfrec}; for image generators,
reconstruction-error methods such as
AEROBLADE~\cite{ricker2024aeroblade} probe whether a diffusion model
reconstructs same-model images more faithfully --- but these rely
on internal latents, not surface-level judgement. To our knowledge,
no prior work has asked a state-of-the-art image generator, in
plain natural language, whether its own document-domain output
looks real, on a paired dataset against authentic counterparts.

%% file: sections_v2/03_dataset_construction.tex
\section{Dataset Construction}
\label{sec:dataset}

\subsection{Source Datasets and Forgery Specifications}
We reuse the four source corpora and the $4{,}062$ forgery specifications
of AIForge-Doc v1~\cite{wu2026aiforgedoc} verbatim:
\textsc{cord} v2~\cite{park2019cord} (1{,}000 Indonesian receipts),
WildReceipt~\cite{sun2021wildreceipt} (1{,}696 English receipts),
SROIE~\cite{huang2019sroie} (946 English receipts), and
XFUND~\cite{xu2022xfund} (420 multilingual forms in seven non-English
languages). Each spec comprises an authentic source image, a target
field, the original textual value, an alternative \emph{forged value}
generated by v1's mutation rules (monetary fields scaled by
$\mathcal{U}(1.15,3.0)$ or $\mathcal{U}(0.20,0.85)$; dates perturbed
within calendar bounds; document IDs digit-flipped), and a pixel
bounding box.

This reuse is deliberate: it makes the v2 forgeries a strict
\emph{paired} extension of v1, so any difference in human or detector
performance between the two datasets is attributable to the generator
(GPT-Image-2 rather than Gemini~2.5 Flash Image / Ideogram~v2 Edit) and
not to the documents being tampered.

\subsection{Generation Pipeline}
\label{sec:pipeline}

\paragraph{Aspect-preserving size snap.}
Our API provider imposes two constraints on the requested output dimensions:
both must be multiples of $16$ in $[16, 3840]$, and their product must
lie in $[655{,}360, 8{,}294{,}400]$ pixels. For each spec we expand
the field bounding box by $50\%$ on each side (minimum $100$~px) to
form a context crop $(W_c, H_c)$, compute the smallest legal
$(W^*, H^*)$ with $W^*/H^* \approx W_c/H_c$ that satisfies both
constraints (e.g.\ $331\times183 \to 1104\times608$), and request that
size. We then \textsc{lanczos}-downsample the returned image to
$(W_c, H_c)$ for downstream paste-back. This is a single
aspect-preserving resize.

The mask, prompt, output-conditioning, and metadata-stripping
details are in Appendix~\ref{app:mask-prompt-output}; the
short version is that we draw a $3$-pixel green-outline marker
(rather than a red fill, which colour-bleeds) on the context
crop, drive the model with a fixed outer wrapper plus a
five-clause inner prompt that pins down character fidelity,
and strip every PNG metadata chunk so judges cannot exploit
provenance leaks.

A $20$-prompt $\times$ $4$-reference prompt-engineering ablation
(\S\ref{app:prompt-ablation-pipeline}) confirms that pipeline
performance is not artefactual: $79/80$ trials produced legible
aspect-correct outputs. Each emitted image then passes a semantic
plausibility check (forged value $\neq$ original; bbox region
differs in pixel space from the authentic counterpart) and an
author-side visual inspection.

%% file: sections_v2/04_statistics.tex
\subsection{Dataset Statistics}
\label{sec:stats}

\paragraph{Scale and composition.}
After the production run, AIForge-Doc v2 contains
\textbf{$3{,}066$} forged document images, each paired with its
authentic source and a pixel-aligned ground-truth mask. The remaining
$996$ specs (24.5\% of the $4{,}062$-spec ceiling) failed deterministic
upstream rejections rather than transient API errors:
$\sim$94\% were \texttt{invalidAspectRatio} (the GPT-Image-2 endpoint
enforces $\text{width}/\text{height} \in [1\!:\!3, 3\!:\!1]$, which
rejects most long-receipt context crops), $\sim$4\% were
\texttt{invalidReferenceImageHeight} (height bounds $[128, 2048]$~px),
and $<$2\% were retry-exhausted timeouts. We treat the remaining
$3{,}066$ as the deployment-relevant dataset.

\begin{table}[t]
\centering
\small
\caption{Source-dataset composition of AIForge-Doc v2. Spec counts
are inherited from v1; \emph{produced} reflects the constraint-driven
acceptance rate of the GPT-Image-2 endpoint, dominated by aspect-ratio
rejection on long receipts.}
\label{tab:scale}
\begin{tabular}{lrrr}
\toprule
Source & Doc.\ type & v1 specs & v2 produced \\
\midrule
CORD v2~\cite{park2019cord}      & Receipt & 1{,}000 & 983 \\
WildReceipt~\cite{sun2021wildreceipt} & Receipt & 1{,}696 & 1{,}336 \\
SROIE~\cite{huang2019sroie}      & Receipt & 946 & 329 \\
XFUND~\cite{xu2022xfund}         & Form    & 420 & 418 \\
\midrule
\textbf{Total}                   &         & \textbf{4{,}062} & \textbf{3{,}066} \\
\bottomrule
\end{tabular}
\end{table}

\paragraph{Field-type distribution.}
Field selection follows v1's priority scheme (financial amount $>$
date $>$ document ID $>$ quantity $>$ other numeric) and is determined
by the source-dataset annotation structure, so the resulting
field-category distribution is identical to v1's.

\paragraph{Mask format and spatial sparsity.}
Ground-truth masks are 8-bit grayscale PNGs at the source resolution.
Pixel value $0$ marks authentic content, $255$ marks the tampered
field bounding box (tight, no padding). The median tampered area is
$5{,}589$~px$^2$, a median of $0.92\%$ of total image pixels
(IQR $[0.35\%, 1.55\%]$): over $99\%$ of pixels in each image are
unmodified, so detection on AIForge-Doc v2 is again a
``needle-in-haystack'' problem at the pixel level --- but as
\S\ref{sec:results} shows, the central result is at the \emph{image}
level.

\paragraph{Cross-generator pairing with v1.}
Because v2 reuses v1's forgery specifications, every successful v2
forged image has a same-spec counterpart produced by v1's Gemini-nano
or qwen-inpaint / Ideogram-v2 Edit pipeline. Of the $3{,}066$
spec\_ids that succeed in v2, $3{,}062$ also have a v1 forgery on
disk ($2{,}729$ Gemini-nano, $333$ qwen-inpaint), holding bounding
box, target value, and source image fixed across generators ---
which supports the per-spec cross-generator comparison reported in
\S\ref{sec:results}.

\subsection{OpenAI Safety Filter}
\label{sec:engineering}

During mass generation we observed a $\sim$$10\%$ rate of
deterministic upstream rejections in which the API provider returns
HTTP~400 with code \texttt{providerInternalError} and the upstream
message \emph{``OpenAI internal error''}. The same
(image, prompt) pair returns the same error across $5$ retries with
up to $300$~s of cumulative backoff, ruling out transient capacity
or rate-limiting and indicating a reproducible upstream rejection.
We interpret this as an undocumented OpenAI safety classifier that
fires on certain document-edit requests.

\paragraph{Distribution of rejected specs.}
The rejections are not uniformly distributed across our spec
catalogue. They cluster sharply on \textbf{financial-amount edits in
the CORD subset of Indonesian retail receipts} --- the spec category
most directly aligned with real-world receipt-fraud (modifying a
\textsc{total}, \textsc{subtotal}, or line-item price). Rejection
rates on field categories that look less obviously fraud-adjacent
(date-only edits, store-address edits, multilingual form fields)
are substantially lower. This pattern suggests OpenAI's policy team
has correctly identified financial-document tampering as a misuse
class worth refusing and shipped a classifier that fires on it.

We expand on the implications for industry coordination, including
the documentation gap and the trivial bypass under prompt
perturbation, in \S\ref{subsec:safety-filter-discussion};
pipeline-level mitigations (retry policy, hard-stop on accumulated
failure) are in Appendix~\ref{app:retry-hardstop}.

%% file: sections_v2/05_judges.tex
\section{Four-Judge Evaluation Protocol}
\label{sec:judges}

We evaluate four judges spanning the practically relevant detection
strategies: human inspectors, a generic forensic detector, the only
published document-specific detector with a released checkpoint, and
the \emph{generator itself} as a zero-shot binary judge. Computational
judges run on the full paired v2 test partition
($n_{\text{forged}} = n_{\text{authentic}} = 3{,}066$); humans on a
$30$-image stratified subset.

\subsection{Human Inspectors}
\label{sec:human}

We collect human judgements through
\href{http://canuspotai.com/}{\texttt{CanUSpotAI.com}}, a public
$2$AFC site we host: each trial shows an authentic--forged pair
from the same spec and asks the visitor which is AI-edited.
$N=120$ non-expert visitors contributed $n=365$ pair-votes on the
GPT-Image-2 receipt subset, reported as a binomial proportion with
a $95\%$ Wilson interval. Balanced $2$AFC accuracy is the
empirical Mann--Whitney $U$ and is directly comparable to the AUC
of the computational judges in Table~\ref{tab:headline}.

\subsection{TruFor: Generic Forensic Detector}
\label{sec:trufor}

TruFor~\cite{guillaro2023trufor} (CVPR 2023) fuses a CLIP-pretrained
ViT-L backbone with NoisePrint++ (a learnable camera-model
fingerprint). It is the strongest published general-purpose forensic
detector on NIST16/Columbia/COVERAGE and achieved AUC $0.751$
zero-shot on AIForge-Doc v1. We use the official checkpoint without
fine-tuning.

\subsection{DocTamper: Document-Specific Detector}
\label{sec:doctamper}

DocTamper~\cite{qu2023doctamper} is the only published detector
trained specifically on document forgeries (Swin Transformer with a
DCT-domain Document Frequency Loss head and a Neighbouring Feature
Coupling module). We use the official qcf-568 checkpoint
(\texttt{dtd\_doctamper.pth}) without fine-tuning. The model emits
a per-pixel forged/authentic mask on $512\times512$ inputs at JPEG
$q\!=\!75$; we aggregate to an image-level score by taking the
fraction of pixels predicted FAKE across the full document.

\subsection{GPT-Image-2 Self-Judge}
\label{sec:self-judge}

The fourth judge is the generator itself, used as a black-box binary
classifier. We query the same \texttt{openai/gpt-5.4-image-2} model
that produced the v2 forgeries via a chat-completions endpoint with
text-only output. The candidate image is submitted with this minimal
prompt:

\begin{quote}\small\itshape
``Is this image AI-edited? Answer with one word: yes or no.''
\end{quote}

The minimal prompt matches the deployment threat model (a black-box
LLM judge in a fraud pipeline) and precludes prompt-engineering
confounds. Each image is queried once. Across $6{,}132$ trials,
refusals and empty completions are zero; $416$ rows ($6.8\%$)
returned text without a single-word yes/no token and are filtered
as ambiguous, with sensitivity bounds in
\S\ref{subsec:prompt-ablation}.

\subsection{Calibration Sets for the Two Forensic Detectors}
\label{sec:calibration}

A chance-level result on AIForge-Doc v2 admits two readings: either
the AI inpainting is invisible to the detector, or the detector is
unable to operate on receipt and form scans regardless of what
tampering is present. To separate the two readings we calibrate
each of the two forensic detectors against a same-source-domain
\emph{traditional} tampering set, constructed to fall inside its
training distribution. If a detector reaches near-published
performance on its calibration set, then its low score on v2
isolates the gap to AI inpainting rather than to a domain-transfer
failure.

\paragraph{TruFor calibration set.}
TruFor's NoisePrint++ residual head depends on CMOS sensor and
in-camera ISP noise being preserved in the input, with cross-camera
splicing as the canonical positive: TruFor reaches AUC $0.996$ on
Columbia and $0.984$ on DSO-1, both natural-photo cross-camera
splicing benchmarks~\cite{guillaro2023trufor}. We replicate this
distribution on our document domain by building $50$ cross-camera
splicing pairs from our source images, pairing each target with a
donor whose JPEG quantisation table differs from the target's (a
proxy for different camera or encoder identity) and hard-pasting a
$128$--$192$\,px patch from the donor into the target without
blending or further re-encoding. Saved as PNG so the sensor-noise
discontinuity at the splice boundary is preserved.

\paragraph{DocTamper calibration set.}
DocTamper's training distribution is OCR-token-aligned tampering on
JPEG-compressed document images, with the CLTD curriculum
recompressing each image $1$--$3$ times at $Q\!\in\![75,100]$. The
model reaches pixel-level F1 $0.99$ on T-SROIE under this
distribution~\cite{qu2023doctamper}. We replicate the
splicing-style positive on our document domain by building $100$
forged--authentic pairs from our source images. Per target image
we replace several numeric OCR tokens (with the count scaled to the
target image's area) with cross-document donor tokens of similar
height and different text, then re-encode the whole image with a
two-pass JPEG cycle ($Q\!=\!85$ then $Q\!=\!75$) to land on the
lower bound of the CLTD curriculum.

The calibration sets carry no overlap of construction details with
v2 (no AI inpainting, no GPT-Image-2-rendered pixels, no spec-list
dependence); they probe \emph{only} whether the detector retains
its published behaviour on our source images when the tampering is
of the type it was trained to detect.

Metrics and statistical treatment (cluster bootstrap,
multiple-comparisons stance) are in Appendix~\ref{app:metrics}.

%% file: sections_v2/06_results.tex
\section{Experiments and Results}
\label{sec:results}

\subsection{Image-level Performance of the Four Judges}

\begin{figure*}[t]
\centering
\includegraphics[width=\textwidth]{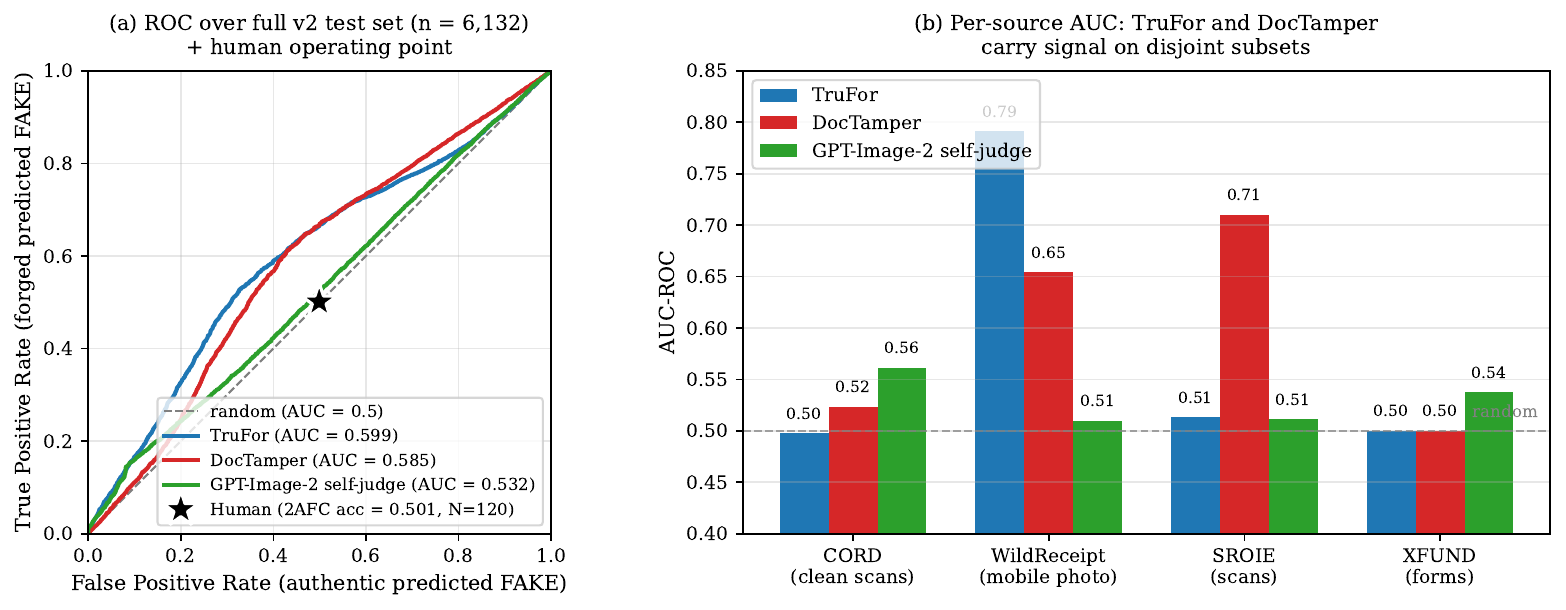}
\caption{(a) ROC curves on the full v2 test set. All three judges
remain close to the chance diagonal: TruFor extracts a small but
statistically significant signal (AUC $0.599$); DocTamper
recovers a comparable signal (AUC $0.585$); the GPT-Image-2
self-judge is statistically indistinguishable from random.
(b) Per-source AUC. TruFor's signal is concentrated in the
\emph{WildReceipt} mobile-photo subset (AUC $0.791$);
DocTamper's signal concentrates in SROIE ($0.710$) and
WildReceipt ($0.654$); both detectors collapse on PDF-rasterised
forms (XFUND).}
\label{fig:roc}
\end{figure*}

Table~\ref{tab:headline} reports the image-level performance of
all four judges on the AIForge-Doc v2 test partition. Human
$2$AFC accuracy is $0.501$ ($n=365$ pair-votes from $N=120$
inspectors via \texttt{CanUSpotAI.com}), statistically
indistinguishable from chance; the three computational judges sit
in $[0.532, 0.599]$. \emph{No judge --- human or computational ---
exceeds AUC $0.6$ in any operationally meaningful sense}. ROC
curves are in Figure~\ref{fig:roc}(a).

\begin{table*}[t]
\centering
\small
\caption{Image-level performance of the four judges on
AIForge-Doc v2. The three computational judges are zero-shot.
Humans were evaluated as a $2$AFC pair-vote task on
\texttt{CanUSpotAI.com} ($N=120$ inspectors, $n=365$ pair-votes
on the GPT-Image-2 subset; \S\ref{sec:human}); the entry shown is
$2$AFC accuracy with a Wilson $95\%$ interval, equal to the
empirical Mann--Whitney AUC. $95\%$ CIs for the computational
judges use $2{,}000$ bootstrap resamples (cluster-by-spec).
TruFor predictions thresholded at $0.5$; DocTamper produces a
per-pixel mask (\S\ref{sec:doctamper}) so binary TPR/TNR at a
fixed threshold are not directly comparable.}
\label{tab:headline}
\setlength{\tabcolsep}{8pt}
\begin{tabular}{lcccc}
\toprule
Judge & AUC & 95\% CI & TPR$_{0.5}$ & TNR$_{0.5}$ \\
\midrule
Random baseline                       & $0.500$ & --- & --- & --- \\
\midrule
Human ($N=120$, $2$AFC)               & $0.501$ & $[0.450,\,0.553]$ & --- & --- \\
TruFor (general forensic)             & $0.599$ & $[0.592,\,0.606]$ & $0.594$ & $0.593$ \\
DocTamper (document forensic, qcf-568)& $0.585$ & $[0.571,\,0.600]$ & --- & --- \\
GPT-Image-2 self-judge (LLM)          & $0.532$ & $[0.525,\,0.537]$ & $0.153$ & $0.910$ \\
\bottomrule
\end{tabular}
\end{table*}

Three observations frame the rest of the section. \textbf{(i)}
The $2$AFC human accuracy of $0.501$ ($95\%$ Wilson interval
$[0.450, 0.553]$) is indistinguishable from random: even non-expert
human inspectors cannot tell GPT-Image-2 receipt forgeries from
their authentic counterparts when shown side by side. The visual
boundary for receipt-domain edits has effectively closed.
\textbf{(ii)} TruFor's $0.599$ is statistically above random
(cluster CI lower bound $0.592$) but achieves only $59.4\%$ TPR at
$40.7\%$ FPR --- close to chance in practice.
\textbf{(iii)} DocTamper produces an above-chance but
operationally inadequate $0.585$ AUC, with the lift concentrated on
the SROIE and WildReceipt subsets (\S\ref{subsec:per-source-detectors}).
The GPT-Image-2 \emph{self-judge} is dominated by the
\textsc{not-AI-edited} reading
($91.0\%$ on authentic, $84.7\%$ on its own forgeries; the $6.3$-pp
gap is the entire signal).

\subsection{Per-Source Behaviour of the Two Forensic Detectors}
\label{subsec:per-source-detectors}

The aggregate TruFor and DocTamper numbers ($0.599$ and $0.585$)
mask substantial per-source variation
(Figure~\ref{fig:roc}(b); detail in
Appendix~\ref{app:metrics}). TruFor's lift is concentrated on
WildReceipt (AUC $0.791$, mobile-camera receipts where the
NoisePrint++ residual has CMOS sensor noise to read), and is
near-random on CORD ($0.498$), SROIE ($0.513$), and XFUND
($0.500$). DocTamper's lift is concentrated on SROIE
($0.710$) and WildReceipt ($0.654$), the two sources whose
JPEG histories are closest to its training-distribution
expectation, and is near-random on CORD ($0.524$, where the
underlying scans already carry strong DCT artefacts that the
detector flags on authentic regions) and XFUND ($0.500$, where
the PDF-rasterised input has no real JPEG history at all).
\emph{The two detectors capture orthogonal subsets of the v2
distribution}: TruFor finds signal where camera noise survives,
DocTamper finds signal where a token-aligned JPEG history
survives, and neither generalises to the rest. Per-image score
correlation between the two is only $\rho=0.31$
(Appendix~\ref{app:judge-correlations}), so a naive averaging
ensemble does not recover the missing coverage.

\subsection{Self-Judge Performance and Prompt Sensitivity}
\label{subsec:prompt-ablation}

\begin{figure*}[t]
\centering
\includegraphics[width=\textwidth]{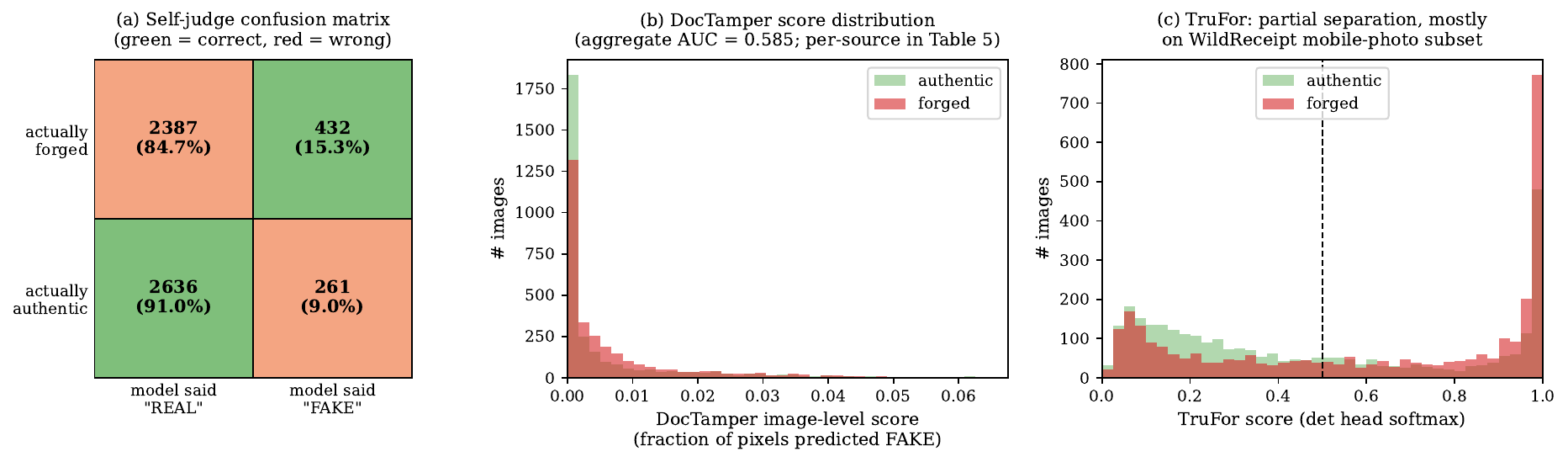}
\caption{Judge biases on AIForge-Doc v2.
(a) Self-judge confusion matrix: model says REAL on $84.7\%$ of
its own forgeries vs.\ $91.0\%$ of authentics; the $6.3$-pp gap is
the entire signal. (b) DocTamper image-level score: forged and
authentic distributions overlap heavily (aggregate AUC $0.585$;
per-source in Table~\ref{tab:per-source}). (c) TruFor's separation
is driven almost entirely by WildReceipt (Figure~\ref{fig:roc}(b)).}
\label{fig:biases}
\end{figure*}

A central finding is that \emph{the same model that
synthesised the forgeries classifies its own output as \textsc{real}
$84.7\%$ of the time}, against a $91.0\%$ REAL rate on authentic
documents (Figure~\ref{fig:biases}(a)). The $6.3$-pp asymmetry is
significant and gives the AUC of $0.532$, but the absolute level is
far from operational.

A natural concern is that this might be an artefact of our
\emph{minimal} prompt --- ``Is this image AI-edited?''
--- and that more elaborate prompting could lift the self-judge out
of the near-random regime. To test this directly we re-judge a
stratified $n_{\text{forged}}=n_{\text{authentic}}=50$ subset under
five prompt variants spanning the strategies most commonly proposed
for LLM-as-judge document forensics: chain-of-thought (P1),
forensic role-play (P2), elaborated AI-edit role-play (P3), and
localisation-hint priming (P4), against the deployment-realistic
minimal baseline (P0). Table~\ref{tab:prompt-ablation} reports AUC
and the forged-vs-authentic FAKE-rate for each.

\begin{table*}[t]
\centering
\small
\setlength{\tabcolsep}{12pt}
\caption{Self-judge prompt-sensitivity ablation
($n_{\text{forged}}=n_{\text{authentic}}=50$). $n_{\text{used}}$ is
the parseable-response count; CoT prompts have lower parse rates
because the model writes multi-paragraph rationales without a
single-word verdict. \emph{No variant lifts AUC above $0.59$.}}
\label{tab:prompt-ablation}
\begin{tabular}{lcccc}
\toprule
Prompt variant & AUC & FAKE\% (forged) & FAKE\% (auth) & $n_{\text{used}}$ \\
\midrule
P0 (minimal, paper)        & $0.522$ & $8.2\%$  & $4.2\%$  & $92$ \\
P1 (chain-of-thought)      & $0.453$ & $18.4\%$ & $25.0\%$ & $64$ \\
P2 (forensic role-play)    & $\mathbf{0.586}$ & $32.7\%$ & $20.8\%$ & $65$ \\
P3 (AI-edit role-play)     & $0.538$ & $4.1\%$  & $0.0\%$  & $50$ \\
P4 (localisation hint)     & $0.529$ & $44.9\%$ & $34.0\%$ & $63$ \\
\bottomrule
\end{tabular}
\end{table*}

Four patterns are notable. \textbf{(i) Chain-of-thought reduces
rather than improves discrimination.} CoT (P1) drops AUC to
$0.453$, indistinguishable from random, and flips FAKE
\emph{more often on authentic} input ($25.0\%$) than on forged
input ($18.4\%$): the reasoning trace amplifies pareidolic
feature-finding rather than detection.
\textbf{(ii) Forensic role-play has the highest AUC but is
operationally infeasible.} P2 gives the best AUC of the five
($0.586$), but at a $20.8\%$ false-positive rate on authentic
inputs --- the forensic frame primes the model to flag normal
document features (faded ink, OCR fonts, wrinkles) as tampering,
so any threshold that yields meaningful TPR gives unacceptable
FPR for a fraud pipeline. \textbf{(iii) Elaborated AI-edit role-play reduces FAKE
commitment.} P3 wraps the AI-edit framing with a multi-clause
forensic context and an explicit reassurance that ``most images
are unedited''; its FAKE-rate on forged input drops to $4.1\%$
(vs.\ P0's $8.2\%$) and to $0.0\%$ on authentic input, so given
an explicit option to attribute edits to AI \emph{plus} a base-rate
prior, the model defaults to REAL. \textbf{(iv) The minimal prompt is not the worst configuration.}
P0 ($0.522$) sits well above P1's chain-of-thought floor
($0.453$) and below the P2 ceiling ($0.586$); the headline
self-judge result is therefore not an artefact of an unusually
pessimistic prompt choice.

Combined with the missing-data sensitivity analysis
(Appendix~\ref{app:ambig-sensitivity}, AUC range $[0.500, 0.568]$
across four fill-in policies for the $6.8\%$ of trials whose
response was not single-word yes/no), the robust statement is:
\emph{no configuration of the self-judge under our protocol
exceeds AUC $\approx 0.59$ on AIForge-Doc v2}. A naive averaging
ensemble does not salvage this --- the self-judge is nearly
orthogonal to TruFor and DocTamper
(Appendix~\ref{app:judge-correlations}).

\subsection{Calibration on Same-Domain Traditional Tampering}
\label{subsec:calibration-results}

We evaluate each forensic detector on its calibration set
(\S\ref{sec:calibration}). Both detectors recover near-published
performance, confirming that the v2 collapse is specific to AI
inpainting rather than to our source domain.

\begin{table*}[t]
\centering
\small
\caption{Image-level AUC ($95\%$ CI) of each forensic detector on
its same-domain traditional-tampering calibration set vs.\ on the
full v2 test partition. \textit{Gap} = AI-inpainting AUC drop.}
\label{tab:calibration}
\setlength{\tabcolsep}{8pt}
\begin{tabular}{lccc}
\toprule
Detector & AUC (calibration) & AUC (v2) & Gap \\
\midrule
TruFor (cross-camera splicing)
                 & $\mathbf{0.962}\;[0.909,\,1.000]$ & $0.599$ & $0.363$ \\
DocTamper (cross-document OCR splicing)
                 & $\mathbf{0.852}\;[0.794,\,0.904]$ & $0.585$ & $0.267$ \\
\bottomrule
\end{tabular}
\end{table*}

TruFor recovers near its published Columbia performance
($0.996$) on the cross-camera splicing calibration set, and
DocTamper clears $0.85$ on the cross-document OCR-token splicing
calibration set; both detectors thus retain near-published
performance on our document domain when the tampering is
traditional. The $0.363$-AUC drop for TruFor and $0.267$-AUC drop
for DocTamper on v2 isolate exactly what AI inpainting removes
that traditional splicing leaves behind: the localised
sensor-noise discontinuity at the PRNU layer, and the localised
JPEG-history mismatch at the DCT layer. AIForge-Doc v2 thereby
quantifies a $0.27$--$0.36$ AUC detection gap that is specific to
AI inpainting on documents and is not attributable to a generic
domain-transfer artefact.

%% file: sections_v2/07_discussion.tex
\section{Discussion}
\label{sec:discussion}

\subsection{Mechanisms of Detector Failure on AI Inpainting}
The two forensic detectors fail on AIForge-Doc v2
(\S\ref{subsec:per-source-detectors}) for a coherent reason that
the calibration results (\S\ref{subsec:calibration-results}) make
explicit: \emph{GPT-Image-2 inpainting removes the localised
discontinuities that today's detectors are trained to find,
without removing them on the rest of the document}.

\paragraph{PRNU channel (TruFor).}
TruFor's NoisePrint++ residual head detects local mismatches in
CMOS sensor and ISP noise. On the calibration set, where the
inserted patch comes from a different camera, this mismatch is
sharp and TruFor reaches AUC $0.962$. On v2, the inpainted
region is rendered fresh by GPT-Image-2 and pasted back into the
original photograph; the inpainted pixels carry no camera
fingerprint at all, but neither do they carry one that
\emph{disagrees} with the surrounding image, because the
generator's per-pixel output noise is itself spatially smooth.
The signal TruFor needs --- a sharp PRNU step at the splice
boundary --- is what AI inpainting most easily removes.

\paragraph{JPEG-history channel (DocTamper).}
DocTamper detects local mismatches in DCT block history at
token-aligned regions. On the calibration set, the spliced
token comes from a different document with its own JPEG history;
the two-pass recompression of the composited image leaves a
residual DCT signature that disagrees with the surrounding
pixels, and DocTamper reaches AUC $0.852$. On v2, by contrast,
GPT-Image-2 produces a single fresh render of the edited region
and the composite is saved as PNG; the edited and surrounding
pixels carry no prior JPEG history at all, so when the detector
ingests the image the resulting DCT field is uniform across the
splice and the history-mismatch signal is gone.

The two channels are independent ($\rho=0.31$;
Appendix~\ref{app:judge-correlations}), so the overlap of
detection on v2 is small and naive averaging does not recover
either detector's calibration ceiling. Future document-forensic
detectors will need signals AI inpainting does \emph{not} make
consistent: typographic micro-features of generated glyphs,
semantic plausibility of the inserted text, or generator-side
provenance signals.

\subsection{Safety-Filter Coverage and the Case for
Industry Coordination}
\label{subsec:safety-filter-discussion}

An incidental but actionable observation from our production run
concerns the deployed safety surface: \emph{OpenAI has shipped
mitigations against misuse, but the existing safety preventions
do not generalise to the document-fraud threat surface}.

\paragraph{Filter coverage and operational limits.}
The $\sim$$10\%$ deterministic safety-filter rejection
(\S\ref{sec:engineering}) concentrates on financial-amount edits
in CORD --- the most fraud-aligned spec category --- so the safety
surface is not absent, but its operational value is limited on
two fronts. First, the classifier is undocumented and its error
surface (\texttt{providerInternalError}) is indistinguishable from
a $5$xx outage. Second, rejection is deterministic on (image,
prompt), so one prompt or bbox perturbation collapses our
$\sim$$10\%$ rejection rate to $0\%$, and the $\sim$$90\%$ that
pass unchallenged are not flagged by any tested detector
(TruFor $0.599$, DocTamper $0.585$, self-judge $0.532$).

\paragraph{Asymmetric awareness and coordination.}
The $0.27$--$0.36$ AUC calibration gap
(\S\ref{subsec:calibration-results}) means universal deployment
of either forensic detector still leaves the AI-document-fraud
surface largely uncovered, and neither generator vendors (who see
prompts but not deployment outcomes) nor forensic vendors (who
see pixels but not refusal signals) close the gap alone. A
coordinated deployment story should include at minimum: published
safety-filter behaviour in model cards, shared red-team datasets
like AIForge-Doc v2 as a common detectability substrate, and
API-level exchange of refusal events or provenance signals
(e.g.\ C2PA content credentials extended to generation gateways).
Ethical considerations are in Appendix~\ref{app:ethics}.

%% file: sections_v2/08_conclusion.tex
\section{Conclusion}
\label{sec:conclusion}

We release \textbf{AIForge-Doc v2}, a paired dataset of $3{,}066$
GPT-Image-2 document forgeries, and benchmark four lines of
defence: humans (\href{http://canuspotai.com/}{\texttt{CanUSpotAI.com}}
$2$AFC, $N\!=\!120$, $n\!=\!365$, $0.501$), TruFor ($0.599$),
DocTamper qcf-568 ($0.585$), and the GPT-Image-2 self-judge
($0.532$). All four sit far below operationally useful.
Calibration on same-domain traditional tampering retains
near-published performance (TruFor $0.962$, DocTamper $0.852$);
switching to GPT-Image-2 inpainting drops AUC by $0.27$--$0.36$,
isolating the gap to AI inpainting. Dataset (CC-BY-4.0) at
\href{https://scam.ai/research}{\texttt{scam.ai/research}}.

%% file: sections_v2/A_appendix.tex
\section{Working Definitions, Scope, and Relation to v1}
\label{app:definitions-scope}

\paragraph{Working definitions.}
This paper is a \emph{measurement study}, not a method proposal:
we report what four judges do on AIForge-Doc v2 and we do not
propose a new detector. We use the following working definitions
throughout. By \textbf{self-recognition} we mean that, under a
fixed natural-language probe, a model assigns higher probability
to the FAKE label on its own forged outputs than on authentic
inputs --- i.e.\ $\Pr(\text{FAKE}\mid x_{\text{forged}}) >
\Pr(\text{FAKE}\mid x_{\text{authentic}})$. By
\textbf{operational utility} we mean the threshold a fraud-screening
pipeline would actually deploy at: AUC $\geq 0.85$ with FPR $\leq 5\%$
at the operating point, taking the published deployment thresholds of
commercial document-forensic services as our anchor; none of our
four judges meets either condition.

\paragraph{Scope.}
Our empirical claims apply specifically to GPT-Image-2 outputs
accessed via public API surfaces; whether they
generalise to other 2026-era image generators is empirically open
and we encourage the test.

\paragraph{Relation to v1.}
AIForge-Doc v1~\cite{wu2026aiforgedoc} established that existing
forgery detectors generalise poorly from Photoshop-style
manipulations to diffusion-based AI inpainting on documents. The
present work is \emph{not} a v1 expansion: it fixes the generator
(GPT-Image-2 only) and asks a different question --- can the
generator itself, queried in plain language, distinguish its own
outputs from authentic counterparts? We treat v1's forgery
specifications as a paired baseline and reuse its source datasets,
so that any difference in detector performance between v1 and v2 is
attributable to generator change rather than dataset change.

\section{Mask, Prompt, and Output Conditioning}
\label{app:mask-prompt-output}

\paragraph{Composite-marker mask.}
GPT-Image-2 does not accept a binary mask channel, so we draw a
marker on the input image that the model is told (via prompt) to
treat as the editing region. v1 used a semi-transparent red
\emph{fill}, which caused chromatic bleed into the generated region
(the forged area was tinged pink in pilot outputs); we replaced it
with a $3$-pixel \emph{green outline}, which the model recognises as
a marker without copying its colour into the synthesised content.

\paragraph{Two-layer prompt.}
Each request carries a fixed outer wrapper that reminds the model
the green rectangle is an overlay marker (must not appear in the
output), and an inner spec-specific layer that pins down character
fidelity in five clauses: (i) declare the requested aspect ratio;
(ii) state the exact target string character-by-character with a
four-character ``tail'' spotlight; (iii) prohibit reformatting (no
inserted spaces/separators); (iv) match the typography of the
original value, optionally anchoring on a longest shared substring;
(v) require all non-mask pixels to remain visually identical. We
arrived at the five-clause structure through ablation
(\S\ref{app:prompt-ablation-pipeline}); shorter prompts under-specify the
character target, and longer ones that cite bounding-box coordinates
led the model to render the literal coordinate text into the image.
The exact prompt template is in our public release.

\paragraph{Metadata stripping.}
Every emitted PNG is rebuilt from raw pixel data into a fresh PIL
\texttt{Image} object so that the file contains no PNG
\texttt{tEXt}/\texttt{iTXt}/\texttt{zTXt}, EXIF, or XMP chunks that
could otherwise leak provenance to the judges. The same stripping is
applied to the matched authentic counterpart so that judges cannot
exploit asymmetric metadata as a shortcut signal.

\paragraph{Generation-asymmetry caveat.}
The forged image passes through one extra processing cycle (rendered
by GPT-Image-2 and then re-encoded as PNG) that the authentic image
does not, which could in principle introduce a low-level spectral
signature. We do not pre-process the authentic image through the
provider to symmetrise the cycle because the model is not
constrained to be the identity. Empirically all three computational
judges fail to extract this asymmetry signal at operationally
useful levels (\S\ref{sec:results}); if a future judge succeeds via
this route, it would be detecting compression history rather than
forgery \emph{per se}, and the appropriate countermeasure is a
uniform re-encoding step at deployment.

\section{Prompt Ablation and Quality Control}
\label{app:prompt-ablation-pipeline}

\paragraph{Twenty-prompt pipeline ablation.}
To verify that pipeline performance is not an artefact of prompt
engineering, we ran a $20$-prompt $\times$ $4$-reference ablation
($80$ trials total): one reference spec from each of the four source
datasets, tested against $20$ prompt variants spanning nine
strategy categories (minimal value-only, short imperative, our
production five-clause prompt, separator hint, character-by-character
spelling, OCR-focused, document-context, role-play
(restoration / Photoshop-retoucher / forensic-examiner),
chain-of-thought, typography expert, colour-control, negative
constraints, forensic-aware, verbose multi-constraint, and a
multi-constraint composite). The full prompt set is in our public
release. Of $80$ trials, $79$ produced legible aspect-correct
outputs. The single failure (CORD reference, ``simple instruction''
variant) returned the same \texttt{providerInternalError} on all
$5$ retries, additional evidence that this rejection is a
deterministic content-policy event rather than transient noise.
Visual grids of all four reference panels are released alongside the
dataset.

\paragraph{Quality control.}
Each emitted image passes a semantic plausibility check (the forged
value differs from the original, and the bbox region in the output
differs in pixel space from the authentic counterpart) and an
author-side visual inspection. An optional PaddleOCR pass was
disabled in production after a NumPy-2.0 incompatibility in the
library; we did not need it, because the visual gate caught the
same failures.

\section{Metrics, Statistical Treatment, and Per-Source Breakdown}
\label{app:metrics}

For each computational judge we report image-level AUC
(area under the ROC curve) with a $95\%$ bootstrap confidence
interval ($2{,}000$ resamples; cluster-by-spec). For TruFor we
additionally report TPR (forged predicted FAKE) and TNR (authentic
predicted REAL) at the $0.5$ decision threshold; for DocTamper,
which emits a per-pixel forged/authentic mask, the natural binary
readout is whether any mask pixels are predicted FAKE, not a
thresholding of the image-level score, so we report AUC only.
For humans we report $2$AFC accuracy on the
\texttt{CanUSpotAI.com} pair-vote stream with a $95\%$ Wilson
binomial interval; on a balanced paired task this is the empirical
Mann--Whitney $U$ statistic and is therefore directly comparable to
the rank-based AUC of the continuous-score
judges.\footnote{The self-judge produces a binary score
(REAL $\to 0$, FAKE $\to 1$). For binary scores, AUC computed via
Mann--Whitney $U$ with tie-handling reduces algebraically to
balanced accuracy $(\text{TPR} + \text{TNR})/2$. We report
self-judge ``AUC'' for notational consistency with the
continuous-score judges, but the self-judge column of
Table~\ref{tab:headline} should be read as balanced accuracy:
$(0.153 + 0.910)/2 = 0.532$. The self-judge AUC is not directly
comparable to the continuous-score AUCs of TruFor and DocTamper as
a rank-ordering measure.} The self-judge has a categorical
output (REAL vs.\ FAKE per trial); we use the per-image FAKE-vote
frequency as the continuous score (with $n_{\text{trials}} = 1$ in
the production run, this collapses to a binary score, but the AUC
computation still distinguishes the two classes via tie-handling).
All confidence intervals are computed against the paired test
partition.

\paragraph{Multiple comparisons.}
Across this paper we report image-level AUC for $4$ judges
$\times$ $4$ sources, a $5$-prompt ablation, and the two
detector calibration sets, in addition to the headline aggregates.
We do \emph{not} apply a Bonferroni or FDR correction across these
subgroup analyses; each is intended as exploratory rather than
confirmatory. The headline judge-vs-random tests on the full paired
test partition (Table~\ref{tab:headline}) are the only inferences we
treat as primary, and their CIs are wide enough to survive any
reasonable correction.

\paragraph{Per-source breakdown.}
Table~\ref{tab:per-source} disaggregates each judge's AUC by source
dataset. The two forensic detectors carry signal on different
subsets of v2: TruFor on the WildReceipt mobile-photo subset
(where camera fingerprints survive), DocTamper on the SROIE and
WildReceipt subsets (where the JPEG history of the original document
is closest to its training-distribution expectation). Both
detectors collapse on XFUND, where the input is PDF-rasterised and
carries neither real camera noise nor a real JPEG history; on
CORD the underlying scans already carry strong DCT artefacts that
DocTamper flags on authentic regions, suppressing the
forged--authentic gap.

\begin{table}[t]
\centering
\small
\setlength{\tabcolsep}{4pt}
\caption{Per-source image-level AUC (\textit{WildR.} = WildReceipt).
The two forensic detectors carry signal on disjoint subsets:
TruFor on WildReceipt only, DocTamper on SROIE and WildReceipt.
Both collapse on XFUND (PDF-rasterised, no real JPEG history).}
\label{tab:per-source}
\begin{tabular}{lcccc}
\toprule
Judge & CORD & WildR. & SROIE & XFUND \\
\midrule
TruFor              & $0.498$ & $\mathbf{0.791}$ & $0.513$ & $0.500$ \\
DocTamper           & $0.524$ & $0.654$          & $\mathbf{0.710}$ & $0.500$ \\
GPT-Image-2 self    & $0.562$ & $0.510$          & $0.512$ & $0.538$ \\
\bottomrule
\end{tabular}
\end{table}

\section{Self-Judge Prompt Compliance and AMBIG Sensitivity}
\label{app:ambig-sensitivity}

\paragraph{Definition of AMBIG.}
The self-judge prompt explicitly asks for a one-word answer:
\emph{``Is this image AI-edited? Answer with one word:
yes or no.''} We parse each response by lower-casing and looking
for a single unambiguous \texttt{yes} or \texttt{no} token. We
label a response \textbf{AMBIG} (ambiguous) when the model writes
text but does not commit to either label --- in practice, a
multi-paragraph rationale (\emph{``This appears to be a receipt with
several indicators that\ldots''}) without ever printing a standalone
\texttt{yes} or \texttt{no}. AMBIG is \emph{not} a refusal and
\emph{not} an empty completion (both occurred zero times in our run);
it simply means the model wrote text but did not follow the one-word
format, so we cannot extract a binary verdict from the response.

\paragraph{Frequency of AMBIG responses.}
Of the $6{,}132$ self-judge trials run on the v2 test set
($3{,}066$ forged $+ 3{,}066$ authentic), $5{,}716$ returned a
parseable yes or no ($93.2\%$ prompt-compliance rate) and
\textbf{$416$ ($6.8\%$) returned an AMBIG response}, splitting
$247$ forged / $169$ authentic. Refusals, empty completions, and
mixed-signal answers (both ``yes'' and ``no'' in the same
response) all occurred zero times.

\paragraph{Effect of AMBIG on the headline AUC.}
The asymmetric split ($247$ vs.\ $169$) raises a
missing-not-at-random concern: the model may be uncertain more often
on harder forgeries, so simply filtering AMBIG rows could bias the
headline AUC. We therefore report the headline two ways: (a) on the
$5{,}716$ parseable rows alone (filter, headline $0.532$), and (b)
under four explicit fill-in policies that bracket the plausible
range (Table~\ref{tab:ambig-sensitivity}). Under the worst case for
the discrimination claim --- forged AMBIG $\to$ REAL \emph{and}
authentic AMBIG $\to$ FAKE --- AUC collapses to exactly $0.500$.
The realistic policy ``model defaults to REAL when uncertain''
(consistent with its overall REAL bias on the parseable rows) gives
$0.528$. Any intermediate probabilistic policy yields an AUC inside
this bracket by linearity of the rank statistic. The substantive
conclusion is unchanged in every case: self-judge AUC $<0.57$ under
\emph{every} fill-in policy.

\begin{table}[t]
\centering
\small
\setlength{\tabcolsep}{4pt}
\caption{Self-judge AUC under four AMBIG fill-in policies. The
headline filtered estimate ($0.532$) lies between the most
conservative bound ($0.500$, exactly random under worst-case
missing-data assumptions) and a hypothetical best-case bound
($0.568$). The plausible realistic policy ``model defaults to REAL
when uncertain'' gives $0.528$.}
\label{tab:ambig-sensitivity}
\begin{tabular}{lc}
\toprule
AMBIG fill-in policy & self-judge AUC \\
\midrule
filter (headline) & $0.532$ \\
all $\to$ REAL & $0.528$ \\
all $\to$ FAKE & $0.541$ \\
worst case (forged $\to$ REAL, auth $\to$ FAKE) & $0.500$ \\
best case (forged $\to$ FAKE, auth $\to$ REAL) & $0.568$ \\
\bottomrule
\end{tabular}
\end{table}

\section{Judge--Judge Score Correlations}
\label{app:judge-correlations}

Pearson correlation of per-image scores across the
$(\text{spec}, \text{kind})$ tuples judged by all three judges:
$\rho(\textsc{TruFor}, \textsc{DocTamper}) = 0.31$,
$\rho(\textsc{Self-judge}, \textsc{TruFor}) = -0.19$,
$\rho(\textsc{Self-judge}, \textsc{DocTamper}) \approx 0$. TruFor and
DocTamper share a mild positive correlation (both target
pixel-level tampering signal but on different channels --- PRNU
residual vs.\ DCT history --- which limits the agreement), while
the GPT-Image-2 self-judge is essentially orthogonal to both. An
ensemble that simply averages the three would not salvage the
result, since the orthogonal directions are roughly random with
respect to the ground truth.

\section{Ethical Considerations and Dual Use}
\label{app:ethics}

This paper documents a measurable gap in current document-fraud
defences and releases the dataset and pipeline that make the gap
reproducible. We outline the dual-use trade-off explicitly.

\paragraph{Rationale for public release.}
The gap already exists in deployed production systems. GPT-Image-2
is generally available at production-grade pricing through at least
three commercial gateways; the underlying composite-marker prompting
technique is already documented in OpenAI's developer community
forum. The marginal capability our release adds for an attacker ---
a curated forgery-spec list with $4{,}062$ entries pre-paired to
authentic source images --- is small relative to the capability the
attacker already has by virtue of the model being on sale. The
marginal capability our release adds for a defender is much larger:
a well-stratified test set, with paired ground-truth masks and
detector calibration sets, that exposes specific failure modes
(PRNU-channel evasion, JPEG-history-channel evasion,
generator-self-recognition collapse) and enables targeted
detector training.

\paragraph{Disclosure timeline.}
We did not pre-disclose the result to OpenAI or to the gateway
providers before submission, because the result describes a
publicly-observable property of an already-shipped model rather than
a new vulnerability. Were OpenAI to issue a mitigation (e.g.\ a
refusal classifier on ``edit a numeric field on a receipt'' prompts),
the v2 generation pipeline would simply stop producing outputs at
deployment time; this would not retroactively invalidate v2 as a
benchmark of pre-mitigation behaviour.

\paragraph{Downstream training.}
The pixel-precise tampered-region masks make AIForge-Doc v2
suitable for fine-tuning document-domain forensic detectors. The
calibration gap (\S\ref{subsec:calibration-results}) suggests
that training detectors on AI-inpainting positives specifically
--- rather than on additional traditional-tampering volume ---
is the highest-leverage follow-on; we explicitly invite that work.

\paragraph{Norms.}
We followed the dual-use disclosure norms of recent forgery-benchmark
releases (DocTamper, OSTF, SAGI~\cite{sagi2025}) and consider the
public release net positive for the defender side of the threat
model.

\section{Retry Policy and Hard-Stop on Failure Rate}
\label{app:retry-hardstop}

\paragraph{Retry policy.}
For non-rejection transient errors (e.g., $504$ Gateway Timeout from
the provider, generic $5$xx upstream from OpenAI) we use $5$ attempts
with exponential backoff $5\,$s, $15\,$s, $45\,$s, $120\,$s, $300\,$s.
This policy is sufficient to resolve essentially all transient
failures while keeping the deterministic safety-filter rejections of
\S\ref{sec:engineering} from inflating wall-clock time.

\paragraph{Hard-stop on accumulated failure rate.}
Generation halts automatically if the running failure fraction
exceeds $1/3$ after at least $10$ specs have been processed, so that
budget cannot be consumed without notice if upstream conditions degrade.
This bound was never reached during the v2 production run.